\documentclass[12pt, letterpaper]{article}
\usepackage[margin=1in]{geometry}

\usepackage[square,,numbers]{natbib}
\usepackage{graphicx,amssymb,amsmath,xcolor}
\usepackage[utf8]{inputenc} 
\usepackage[T1]{fontenc}    
\usepackage{hyperref}       
\usepackage{url}            
\usepackage{booktabs}       
\usepackage{amsfonts}       
\usepackage{nicefrac}       
\usepackage{microtype}      

\usepackage{graphicx,amssymb,amsmath,xcolor,amsthm,url,multirow,array,nicefrac,comment,subcaption,cleveref}

\newcommand{\tabincell}[2]{\begin{tabular}{@{}#1@{}}#2\end{tabular}}

\renewcommand{\cite}[1]{\citep{#1}}
\renewcommand{\citet}[1]{\citep{#1}}

\def\bbE{{\mathbb E}}
\def\bbR{{\mathbb R}}

\newcommand*{\affaddr}[1]{#1} 

\newcommand*{\email}[1]{\texttt{#1}}

 \title{Exploring Model Robustness with Adaptive Networks and Improved Adversarial Training}

\author{
Zheng Xu, \, Ali Shafahi \, Tom Goldstein \\
\email{xuzh,ashafahi,tomg@cs.umd.edu} \\
\affaddr{University of Maryland, College Park}\\
}
\date{}

\begin{document}
\sloppy

\maketitle

\begin{abstract}
Adversarial training has proven to be effective in hardening networks against adversarial examples. 
However, the gained robustness is limited by network capacity and number of training samples.
Consequently, to build more robust models, it is common practice to train on widened networks with more parameters. 
To boost robustness, we propose a conditional normalization module to adapt networks when conditioned on input samples.
Our adaptive networks, once adversarially trained, can outperform their non-adaptive counterparts on both clean validation accuracy and robustness. Our method is objective agnostic and consistently improves both the conventional adversarial training objective and the TRADES objective. 
Our adaptive networks also outperform larger widened non-adaptive architectures that have 1.5 times more parameters.
We further introduce several practical ``tricks'' in adversarial training to improve robustness and empirically verify their efficiency. 
\end{abstract}

\section{Introduction}
Deep neural networks have achieved impressive performance on many machine learning tasks, which has led to growing interests in deploying these models in practical applications. 
However, recent studies have revealed that models trained on benign examples are susceptible to {\em adversarial examples}, examples crafted by an adversary to control model behavior at test time \citep{biggio2013evasion,szegedy2013intriguing, goodfellow2014explaining}. The adversarial perturbation overlaid on top of the benign examples is often small enough to be imperceptible to humans, yet can cause the model to misclassify the image. 

The existence of adversarial examples has raised security concerns for many high-stakes real-world applications such as street sign detection for autonomous vehicles.  
While initial works stated that digital adversarial examples  built for sign-detection may not be a real threat since the camera can view the objects from different distances and angles~\citep{lu2017no}, more recent attacks were proposed for making stronger adversarial examples that are invariant to various transformations by optimizing over the expected value of a set of pre-defined  transformations~\citep{athalye2017synthesizing}. In fact, this security concern has turned into an actual threat after a recent study showed that adversarial stickers are able to fool real-world self-driving cars~\citep{Tesla2019}. 
These security concerns and threats have guided researchers to create models that are both accurate in prediction and robust to attacks.

Various methods have been proposed for defending against adversarial examples. One popular approach is to detect and reject adversarial examples \citep{ma2018characterizing,meng2017magnet,xu2017feature}, which can be ineffective when the adversary is aware of the detection method in order to adapt accordingly \citep{carlini2017adversarial}. Another approach is to introduce regularization for training robust models \citep{CisseBGDU17,jakubovitz2018improving}, but the increase in robustness from such methods is limited.
\citet{athalye2018obfuscated} showed that many proposed defenses give a false sense of security by obfuscating gradients, as meaningful gradient information is necessary for optimization based attacks. 
\citet{athalye2018obfuscated} broke these defenses by attacks that build good approximations for the gradients.  
Among various defense methods, adversarial training \citep{madry2017towards,kannan2018adversarial,xie2018feature,shafahi2018universal} is one of the most common methods for training robust models. In adversarial training, a robust model is trained on adversarial examples that are generated on-the-fly, which is effective but also makes adversarial training expensive. 

Robust models have some interesting properties that have been revealed in recent studies. First, it is argued that there exists trade-offs between accuracy and robustness \citep{tsipras2018robustness,zhang2019theoretically,su2018robustness}. It is difficult to make a model robust to all samples while maintaining the same level of accuracy. 
Second, it is difficult to adversarially train robust models that generalize since adversarially robust generalization requires more data \citep{schmidt2018adversarially} and models with more capacity \citep{madry2017towards}. Training high capacity models on large datasets increases the cost of adversarially training robust models. 
Third, while adversarial training is expensive, it is shown that adversarially trained models learn feature representations that align well with human perception \cite{tsipras2018robustness}.
These feature embeddings can produce clean inter-class interpolations similar to generative models in Generative Adversarial Networks (GANs)~\citep{goodfellow2014generative}. These properties have inspired us to explore model capacity and sample efficiency. 

Recently, conditional normalization, built upon instance normalization~\citep{ulyanov2016instance} or batch normalization~\cite{ioffe2015batch}, has been successful in generative models~\cite{karras2018style} and style transfer~\cite{huang2017arbitrary}. Conditional normalization can be seen as an adaptive network that shifts the statistics of a layer’s activations by applying network parameters conditioned on the latent factors such as style and classes \citep{de2017modulating,dumoulin2016learned}. 
Inspired by these studies, we propose to exploit adaptive networks for robustness in the adversarial training framework. 

\subsection*{Contributions}

We propose building hardened networks by adversarially training {\em adaptive networks}.
To build adaptive networks, we introduce a normalization module conditioned on {\em inputs} which allows the network to ``adapt'' itself for different samples. 
The conditional normalization module includes a meta-convolutional network that changes the scale and bias parameter for normalization based on input samples. 
Conditional normalization is a powerful module that enlarges the representative capacity of networks.
Our adversarially trained adaptive nets can be potentially more robust than conventional non-adaptive nets as they can adapt the network to be robust to adversarial attacks on a specific sample instead of all samples. 
Furthermore, adaptive normalization adds far fewer parameters than other methods for increasing expressiveness and robustness (e.g. wide resnets).


Our experiments on the CIFAR-10 and CIFAR-100 benchmarks empirically show that our proposed adaptive networks are better than their non-adaptive counterparts. 
The adaptive networks even outperform larger networks with more parameters in terms of both (clean) validation accuracy and robustness. 
Moreover, we have made several key observations that not only help our understanding but also significantly boost the performance of adversarial training.
Such ``tricks'' like larger step-size and initializing with a natural model can be widely used in adversarial training, and help us build stronger baselines for non-adaptive networks. 
Our adaptive network outperforms previous reported results by about 4\%, and the strong baselines we achieved by about 1\% in robust accuracy.

The proposed adaptive network can be combined with various other methods to improve the robustness against adversarial examples. Besides extensive experiments with our improved fast adversarial training, we show adaptive networks can be combined with the stronger TRADES~\citep{zhang2019theoretically} objective formulation that is very effective for the CIFAR benchmark, which suggests that our method is objective agnostic and can be helpful in improving many of the well-established baselines.  
Finally, we introduce a variant of single-step adversarial training, when combined with adaptive network, can approach the robust accuracy of non-adaptive network with multi-step adversarial training. 
Though our single-step variant performs slightly worse than our improved fast adversarial training, it complements recent interests in accelerating adversarial training and showcases why conventional single-step methods did not result in robustness against iterative attacks. 

\section{Related work}
Here we provide a brief overview of robustness and normalization layers which are closely related to our proposed adaptive networks. We also provide an overview of adversarial training, which plays a critical role in our method, 

\vspace{0.2cm}
\noindent\textbf{Robustness}, in the white-box threat model, is commonly measured by computing the accuracy of the model on adversarial examples constructed by gradient-based optimization methods starting from validation samples. 
This evaluation method provides an upper-bound on robustness as there is no theoretical guarantee (at least for all classes of problems) that adversarial examples crafted using first-order gradient information are optimal. 
From a theoretical point of view, finding optimal adversarial examples is difficult. Some recent works have proposed finding the optimal solution by modeling neural networks as Mixed Integer Programs (MIPs) and solving those MIPs using commercial solvers~\citep{tjeng2017evaluating}. However, finding the optimal solution of an MIP is generally NP-hard. Although recent advancements have been made in their formulations by enforcing some properties on the network~\citep{xiao2019training}, finding the optimal solution is only feasible for small networks and is very time consuming. That is why certified methods in practice provide lower-bounds on the size of perturbations needed for causing misclassification by solving a relaxed version of the problem.

\citet{raghunathan2018certified} propose certified defences by including a differentiable certificate as a regularizer. Many studies follow this line of work and propose certified defenses~\citep{wang2018mixtrain,wong2018scaling,cohen2019certified}. While from a theoretical point of view certified defenses are interesting, in practice, adversarial training is still the most popular method for hardening networks -- leaders of various computer vision defense competitions and benchmarks utilize adversarial training in their approach \citep{xie2018feature, zhang2019theoretically, madry2017towards}. 

\vspace{0.2cm}
\noindent\textbf{Adversarial training}, in its general form, corresponds to training on the following loss function,
\begin{equation}
    \underset{\theta}{\mathrm{min}} \,
    \sum_i  \kappa J(f_\theta(x_i), y_i) + (1-\kappa) J(f_\theta(x_i+\delta_i), y_i)
\label{eq:adv_training}
\end{equation}
where $J$ is a differentiable surrogate loss used for training the neural network such as the cross-entropy loss, ($x_i, y_i$) is the $i^\text{th}$ data-point and its correct label,  $f_\theta$ is the network with  trainable parameters $\theta$, $\kappa$ is a hyper-parameter that controls how much weight should be given to training on natural examples, and $\delta_i$ corresponds to the adversarial perturbation for the $i^\text{th}$ sample. To keep the perturbation unrecognizable to humans, the norm of $\delta_i$ is often bounded. Throughout this paper, we will use the common $\ell_\infty$-norm bound on $\delta$. Note that our adversarial training loss merges information from both natural and adversarial examples. 

Early adversarial example generation methods required many iterations since their goal was to help an attacker build an adversarial example using minimal perturbations~\citep{szegedy2013intriguing, moosavi2016deepfool, carlini2017towards}. However, from a defender's perspective, the goal is to train on {\em fast} and bounded adversarial examples. 
With speed in mind, \citet{goodfellow2014explaining} proposed training on a single-step $\ell_\infty$ attack called the Fast Gradient Sign Method (FGSM). 
FGSM computes $\nabla_x J(x,y,\theta)$ and sets $\delta = \epsilon \cdot \text{sign}(\nabla_x J(x,y,\theta))$, where $\epsilon$ is the perturbation bound. 
Later, it was shown that stronger attacks such as BIM \citep{kurakin2016adversarialBIM}, completely break FGSM adversarially trained models. 
The BIM attack can be seen as an iterative version of FGSM where during each iteration, the perturbation is updated using an FGSM-type step but with a step-size $\epsilon_s$ which is usually smaller than $\epsilon$,
\begin{equation}
    \delta^k = \delta^{k-1} + \epsilon_s \cdot \text{sign}(\nabla_{\delta} J(f_\theta(x+\delta^{k-1}), y))
    \label{eq:pgd_update_step}
\end{equation}
where $\delta^k$ is the perturbation at iteration $k$ of the BIM attack. After every iteration of the BIM attack (equation~\eqref{eq:pgd_update_step}), $\delta^k$ is clipped such that $\delta^k\in[-\epsilon, \epsilon]$.

Adversarial training started blooming when \citet{madry2017towards} proposed training on adversarial examples generated using the PGD attack, which is a variant of BIM with a random initialization and projection back on the $\ell_p$-norm ball.  
Through experiments, they showed that the PGD attack is the strongest first-order adversary, which was later verified by \citet{athalye2018obfuscated}. Consequently, almost all of the successful adversarially trained robust models use the PGD algorithm to generate adversarial examples.


Training on adversarial examples generated using PGD increases the cost of training by a factor of $K$, where $K$ is the number of iterations of the PGD attack (i.e., number of times we update $\delta$ using equation~\eqref{eq:pgd_update_step}). While we will use PGD-K attacks for evaluating the robustness of all our models, due to the high computation cost associated with PGD adversarial training, we perform most of our adversarial training by modifying a recently proposed algorithm for speeding up adversarial training~\citep{shafahi2018free}. 
A recent study \citep{anonymous2019} suggested a well tuned single-step adversarial training can defend against strong PGD adversarial examples. 
However, the method in \citep{anonymous2019} heavily depends on domain specific cyclic learning rate schedule, using a step-size $\epsilon_s$ which is greater than $\epsilon$, and early stopping based on frequent examination. Also, they only justify their results empirically without providing intuition on why this rather unconventional setup ($\epsilon_s > \epsilon$) is needed. 

\vspace{0.2cm}
\noindent\textbf{Normalization layers} such as batch normalization \cite{ioffe2015batch} and instance normalization \cite{ulyanov2016instance} have become important modules in modern neural networks. Normalization layers standardize input to have zero mean and unit variance, and then shift these statistics using scaling and bias parameters.  \citet{zhang2019fixup} suggest scaling and bias parameters can be even more important than standardization. Conditional normalization, where scaling and bias are adaptively determined by latent factors, has shown to be powerful in many computer vision tasks including style transfer \cite{huang2017arbitrary,dumoulin2016learned} and generative adversarial networks \cite{karras2018style}.

\section{Adaptive Networks}
We introduce adaptive networks with conditional normalization modules in this section. 
Our motivation for adding conditional normalization modules is two-fold. First, by introducing adaptive layers conditioned on inputs, we can ``adapt'' a trained network to be more robust to an individual input sample without requiring any information about its class label -- a useful trait for robustness evaluation.

Second, conditional normalization can increase the expressiveness and effective capacity of the network, which has been shown to have a positive effect in improving model robustness. Adversarially trained models with more expressive capacities are more robust than their less expressive alternatives~\citep{madry2017towards,shafahi2018free}. At a high level, these conditional normalization modules can be considered as adding multi-branch structures to a network which is known to be effective in improving accuracy on validation examples~\cite{huang2017densely}. 
As we will see in the experiments, our normalization module indeed does improve the clean validation accuracy and is more effective\footnote{The adversarially trained adaptive nets have higher validation accuracy and robustness compared to networks with more trainable parameters.} than simply widening or concatenating features in practice.

Below, we show how to create an adaptive network by adding conditional normalization modules to the wide residual network (WRN) \cite{zagoruyko2016wide} architecture.

\begin{figure}[t]
    \centering
    \includegraphics[width=.7\linewidth]{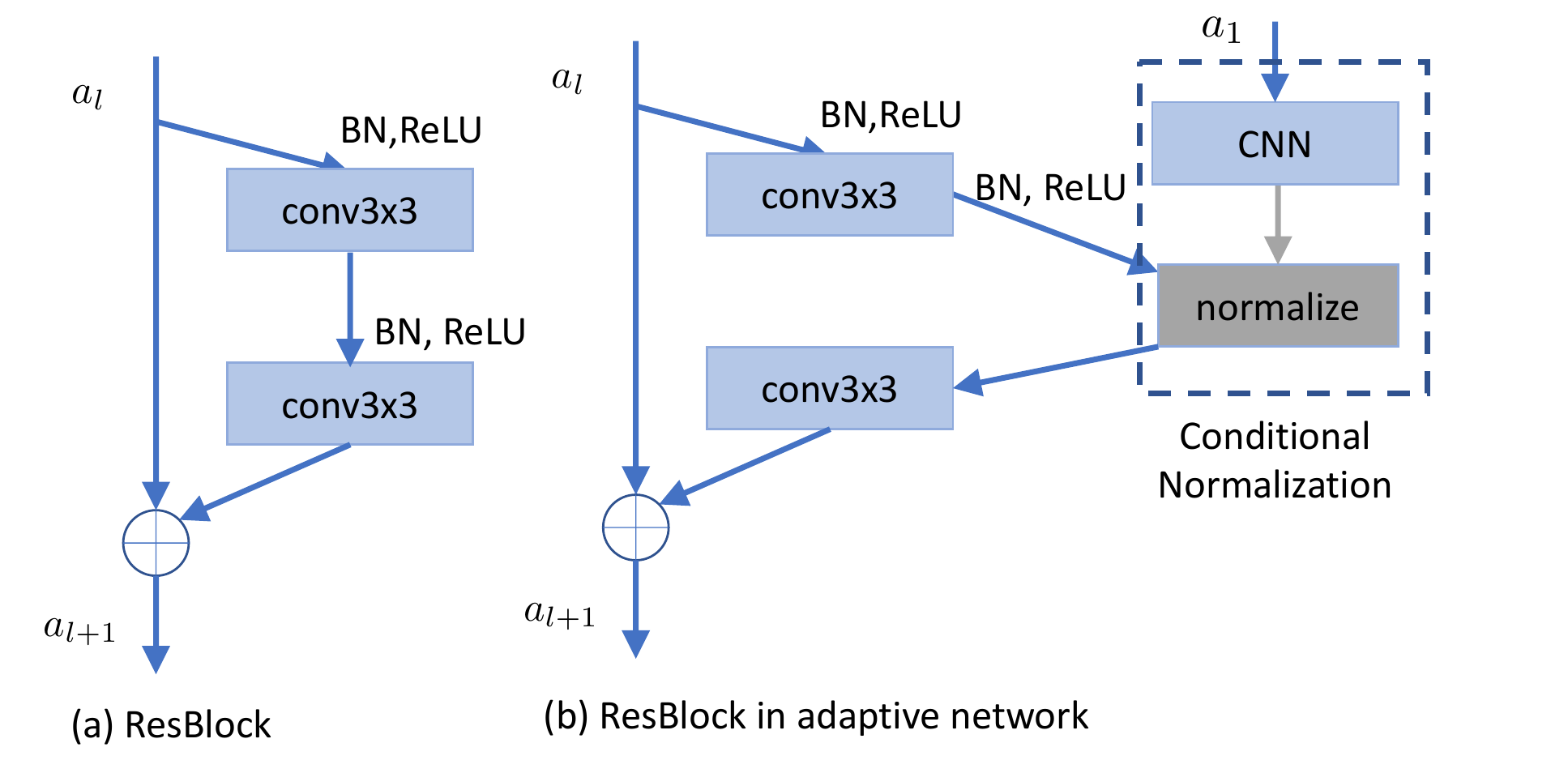}
    \caption{
    Network architecture with adaptive layers. 
    }
    \label{fig:net}
\end{figure}

\subsection{Network architecture}
Let $x\in \bbR^{N\times C\times H\times W}$ represent the feature maps of a convolutional layer for a minibatch of samples, where $N$ is the batch size, $C$ is the width of the layer (number of channels), and $H$ and $W$ are the feature map's height and width. If $x_{nchw}$ denotes the element at height $h$, width $w$ of the $c^\text{th}$ channel from the $n^\text{th}$ sample, the conditional normalization module transforms the feature maps as,
\begin{equation}
\begin{split}
\text{Norm}(x_{nchw} | z) & = \nu(z)_{nc} \, x_{nchw} + \mu(z)_{nc}, \label{eq:norm}
\end{split}
\end{equation}
where $\nu(z),\mu(z) \in \bbR^{N\times C}$ are scale and bias parameters of the normalization module. 
The network with conditional normalization becomes adaptive to the latent factor $z$ as $\nu(z),\mu(z)$ are outputs of convolutional networks with trainable parameters.  
Equation~\eqref{eq:norm} represents normalization in a general form: when latent factor $z$ is a style image and $x$ is normalized by its mean and variance, equation~\eqref{eq:norm} becomes adaptive instance normalization for image style transfer \cite{huang2017arbitrary}; when latent factor $z$ is latent code like random noise, equation~\eqref{eq:norm} becomes the building module for the generator in StyleGAN \cite{karras2018style}. We provide details on how we use input sample as latent factor $z$ as below. 

In our experiments, we add our conditional normalization module to wide residual networks (WRNs) \cite{zagoruyko2016wide} to create adaptive networks for classification. 
WRNs are a derivative of ResNets~\cite{He2015}, and are one of the state-of-the-art architectures used for image classification. 
A WRN is a stack of residual blocks (\cref{fig:net}~(a)). To specify WRNs, we follow~\cite{zagoruyko2016wide} and denote the architecture as WRN-$\beta$-$\alpha$, where $\beta$ represents the depth and $\alpha$ represents the widening factor of the network.

The WRN architecture for the CIFAR-10 and CIFAR-100 datasets we use in this paper consists of a stack of three groups of residual blocks. There is a downsampling layer between two groups, and the number of channels (width of a convolutional layer) is doubled after downsampling.  In the three groups, the width of the convolutional layers are $\{16 \alpha, 32 \alpha, 64 \alpha\}$, respectively. 
Each group contains $\beta_r$ residual blocks, and each residual block contains two $3 \times 3$ convolutional layers equipped with ReLU activation and batch normalization. 
There is a $3 \times 3$ convolutional layer with $16$ channels before the three groups of residual blocks.
And there is a global average pooling, a fully-connected layer and a softmax layer after the three groups. 
The depth of the WRN is $\beta = 6 \beta_r + 4$. 

We add conditional normalization for the first residual block of each of the three groups. The normalization module is applied between the two convolutional layers in a block, as shown in \cref{fig:net}~(b). The inputs to the conditional normalization module are the feature maps produced by the first convolutional layer. Our conditional normalization module consists of a three layer convolutional network: two $3\times 3$ convolutional layers with $16 \alpha$, and one $1 \times 1$ convolutional layer to match the dimension of the three different residual blocks, $2\times \{16 \alpha, 32 \alpha, 64 \alpha\}$, respectively. We use average pooling as the last layer to get $\nu(z),\mu(z)$ for equation~\eqref{eq:norm}. 
Our adaptive network is only slightly larger than the original WRN, and becomes more robust when adversarially trained, as shown in \cref{sec:exp}.

\section{Adversarial training}\label{sec:adv_trains_we_use}
We briefly review the adversarial training algorithm we will use to make our adaptive networks robust, and discuss the ``tricks'' we found useful in improving these algorithms. We then introduce a variant of single-step adversarial training that can couple with standard natural training without extra tuning, and shed light on why this single-step method works and why the conventional single-step methods fail to become robust against PGD attacks. Finally, we discuss an alternative objective function for adversarial training as adaptive networks can complement other active research directions in improving the practical robustness of networks.

\noindent\textbf{PGD adversarial training.}  Well-known robust networks on MNIST and CIFAR-10 were adversarially trained by \citet{madry2017towards} by setting $\kappa=0$ in equation~\eqref{eq:adv_training} and only training on adversarial examples. 
Training just on adversarial examples is justified from a robust optimization framework, and modeled as a two-player constant sum game between the adversary, which is in charge of the perturbation $\delta,$ and the classifier with network parameters $\theta$. 
Formally, we consider adversarial training based on the following minimax formulation, 
\begin{equation}
    \underset{\theta}{\mathrm{min}} \,\, \underset{\delta_i}{\mathrm{max}} \,
    \sum_i  J(f_\theta(x_i+\delta_i), y_i) \,\
    \text{s.t.} \,\ \|\delta_i\|_{\infty} \leq \epsilon \,\ \forall i
    \label{eq:madry_adv_train}
\end{equation}
\citet{madry2017towards} solved the optimization problem in equation~\eqref{eq:madry_adv_train} in an alternating fashion. Before each minimization step on the network parameters $\theta$, they compute $\delta$ using a PGD-K attack on the fly. 
Every perturbation update step of the PGD-K attack (equation~\eqref{eq:pgd_update_step}) requires computing $\nabla_{\delta} J(f_{\theta^j}(x_j+\delta^{k-1}_j), y_j)$, where $\delta^{k-1}_j$ are the adversarial perturbations of the j$^\text{th}$ mini-batch after the previous $k-1$ times $\delta$ update step, and $\theta^j$ represents network parameters at the j$^\text{th}$ minimization iteration. To compute $\nabla_{\delta} J(f_\theta(x_i+\delta^{k-1}_i), y_i)$, required for every step of PGD-$K$, we need a complete forward and backward pass on the network. As a result, every iteration of PGD adversarial training is $(K+1)$ times more expensive than an iteration of natural training. A typical value used for $K$ is 7 to train a robust model for CIFAR-10 benchmark \citep{madry2017towards}.

\vspace{0.2cm}
\noindent\textbf{Fast adversarial training.} To speed up training of robust models,
we adopt a fast adversarial training algorithm recently proposed by~\citet{shafahi2018free}. \citet{shafahi2018free} showed that they can achieve comparable robustness to PGD adversarial training \citep{madry2017towards} on the datasets of our interest (CIFAR-10 and CIFAR-100) with roughly the same cost as standard (non-robust) training.

The fast algorithm (Free-$m$) has a perturbation parameter $\delta_b$ of shape $N\times C\times H\times W$ which is updated once during every minimization iteration. To accelerate robust training, Free-$m$ applies simultaneous updates for the network parameters $\theta$ and perturbation $\delta$, which makes its computation cost almost the same as natural training. In the j$^\text{th}$ minimization iteration, both $\nabla_{\delta}J$ and $\nabla_{\theta}J$ are computed for the current mini-batch $(x_j, y_j)$ and network parameters $\theta^j$,
\begin{align*}
    \nabla_\theta J &= \bbE_{\{(x_j,y_j)\}} [\nabla_\theta \, J(f_{\theta^j}(x_j+\delta^j_b), y_j)] \\
	\nabla_\delta J &=  \nabla_{\delta} \, J(f_{\theta^j}(x_j+\delta^j_b), y_j)]
\end{align*}
Then $\theta$ and $\delta$ are updated as,
\begin{align*}
        &\theta^{j+1} = \theta^j - \tau \nabla_\theta J \\
    	&\delta^{j+1}_b = \text{clip} (\delta^j_b + \epsilon_s \cdot\text{sign}(\nabla_x J), -\epsilon, \epsilon).
\end{align*}
In Free-$m$, each mini-batch is replayed $m$ times. For example, if $m=2$, we move on to the next mini-batch every other step, and therefore the data for the first two iterations would be the same (i.e., $(x_1, y_1)=(x_{2}, y_{2})$). 
Since we train on the same mini-batch $m$-times in a row, the hyper-parameter $m$ is more-or-less analogous to the number of iterations of the PGD training algorithm $K$.
We use the same number of minibatch updates for Free-$m$ adversarial training and natural training on clean images, i.e., we train Free-$m$ for $1/m$ number of epochs in total. 
Free-$m$ can achieve similar robustness accuracy as PGD-$K$ adversarially trained models. In our modification, we apply two ``tricks'' which we found to be particularly effective when combined with free adversarial training: initializing with the natural trained model and applying larger step-size for updating perturbations. We built stronger baselines with such techniques, which can be even further boosted with our adaptive networks.

\begin{figure}[tbhp]
    \centering
    \begin{subfigure}{.49\linewidth}
    \includegraphics[width=\linewidth]{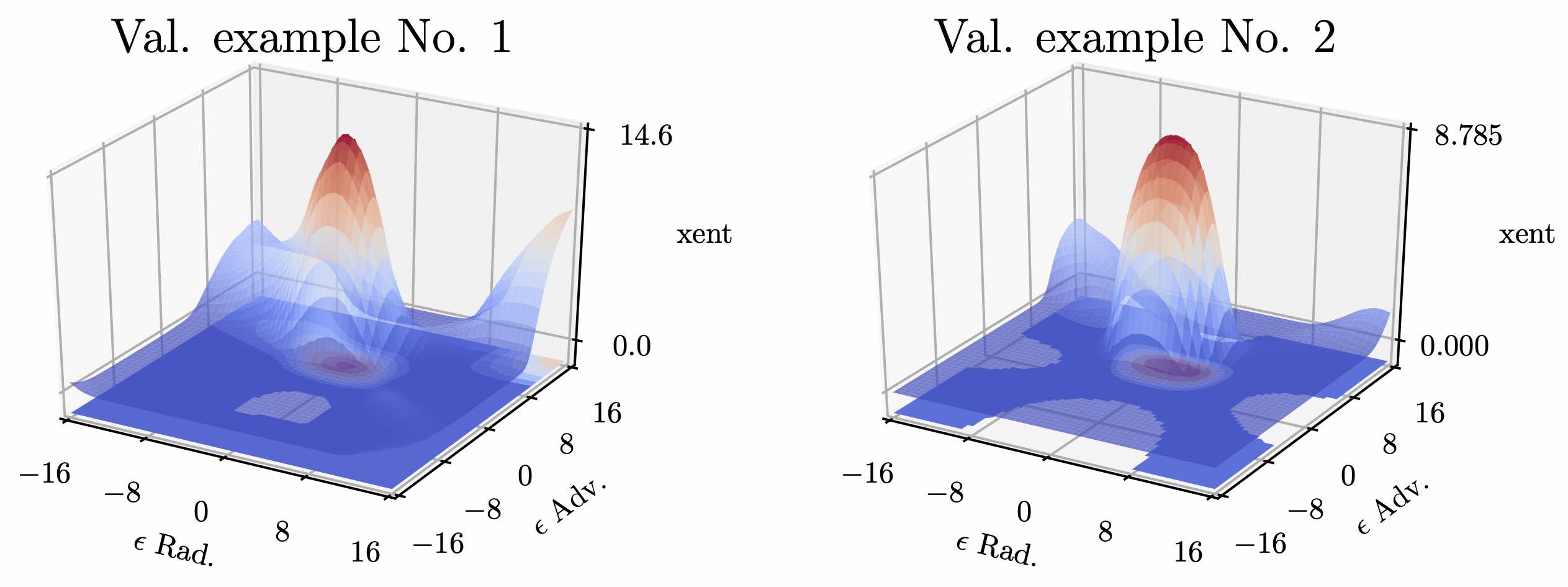}
    \caption{Classical RFGSM}
    \end{subfigure}
    \begin{subfigure}{.49\linewidth}
    \includegraphics[width=\linewidth]{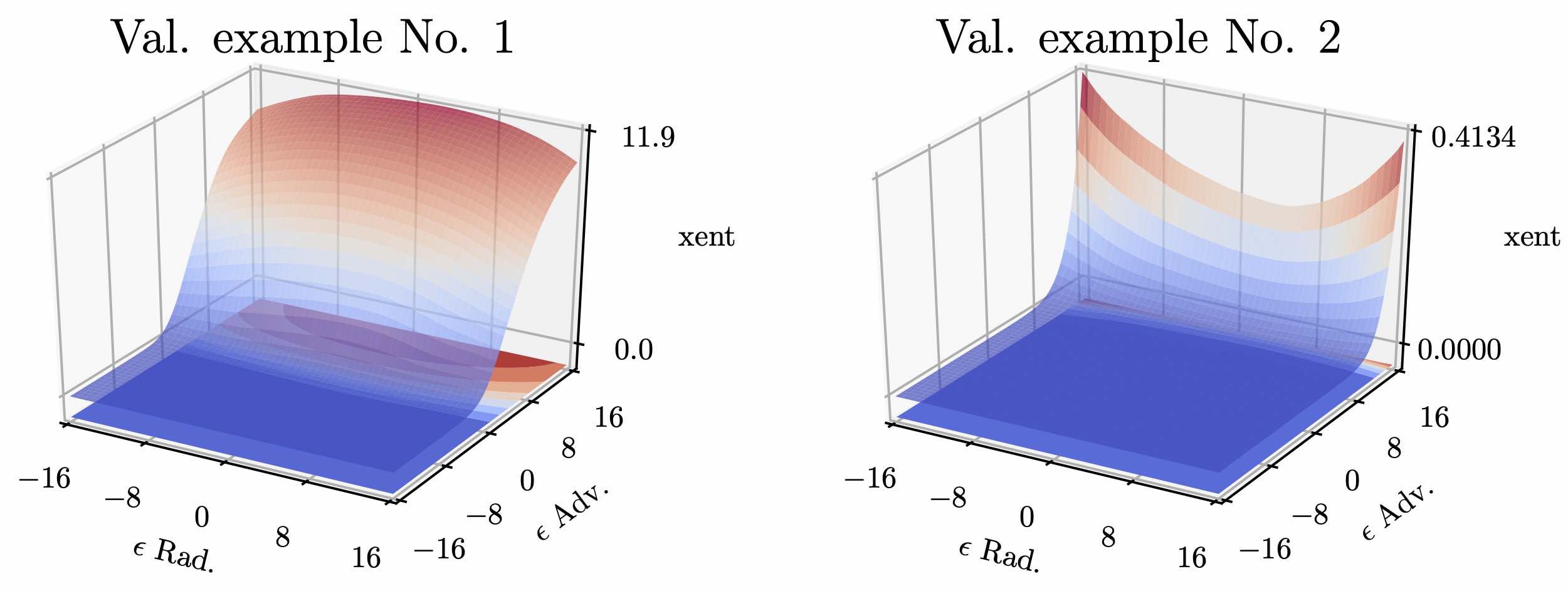}
    \caption{\text{Proposed RFGSM}}
    \end{subfigure}
    \caption{Loss surface plots for points surrounding the first two CIFAR-10 validation images for models adversarially trained on 
    (a) classical RFGSM and (b) proposed RFGSM. The proposed method generates smoother surface, even for a misclassified example (No. 1). 
    }
    \label{fig:loss_surface_FGSMs}
\end{figure}

\vspace{0.2cm}
\noindent\textbf{Single-step adversarial training.} 
Fast Gradient Sign Method (FGSM) \citep{goodfellow2014explaining} is one of the most popular single step method for generating adversarial examples. 
With a random initialization of perturbation, Random FGSM (RFGSM) is similar to doing a one step of the PGD algorithm.
\citet{madry2017towards} showed robust model adversarially trained with FGSM and RFGSM have almost zero robust accuracy under PGD attacks. 
A more recent preprint \citep{anonymous2019} suggested RFGSM-based training can be used to defend PGD attacks when combined with cyclic learning rates and early stopping by examining the robust accuracy on the validation dataset. 
The RFGSM method in \citep{anonymous2019} provides an alternative way to train robust models besides PGD adversarial training \citep{madry2017towards} and fast adversarial training \citep{shafahi2018free} on benchmark datasets such as CIFARs. 
However, it may encounter difficulty to generalize to problems without special learning rate schedules and problems where we cannot perform online validation for early stopping. 

We introduce a variant of RFGSM that works well even with a normal training schedule. We make two key modifications to the classical RFGSM. First, instead of initializing from uniform random value between $-\epsilon$ and $\epsilon$, we initialize from a normal distribution with zero mean and $\sigma^2$ variance. We find $\sigma$ to be rather insensitive between $\epsilon$  and $3\epsilon$, and always use $2\epsilon$ in experiments. Second, we do not clip the perturbation after the FGSM update. Note that the perturbation is still bounded to some extent as the stepsize of FGSM is $\epsilon$. In the proposed variant, the initialized noise can be viewed as boosting training samples instead of the FGSM update.

The classical RFGSM may fail because adversarial examples generated by FGSM during training are likely to fall on the boundary of the $\epsilon$ bounded $\ell_p$ ball. After training on those adversarial examples, the loss surface becomes smooth at the boundary but the cross-entropy loss may take on large values within the $\epsilon-\ell_p$ ball which can be exploited by multi-step methods like PGD. As shown in \cref{fig:loss_surface_FGSMs}, the proposed RFGSM makes the loss surface smoother and hence harder to attack. Even for a difficult sample (validation example id 1), where there are adversarial examples for models trained by both the classical RFGSM and the proposed RFGSM, the loss surface of our proposed RFGSM is smoother. 

\vspace{0.2cm}
\noindent\textbf{TRADES objective.} The proposed adaptive network is complementary to the choice of objective in adversarial training. Besides the minimax problem in equation~\eqref{eq:madry_adv_train}, we can also train adaptive networks with the TRADES objective proposed in \cite{zhang2019theoretically}. TRADES achieves impressive robust accuracy on the CIFAR-10 dataset by combining supervised training and virtual adversarial training as,
\begin{equation}
\small
\begin{split}
    \underset{\theta}{\mathrm{min}} \,\, \underset{\delta_i}{\mathrm{max}} \, &
    \sum_i  J(f_\theta(x_i), y_i) + \frac{1}{\lambda} J(f_\theta(x_i+\delta_i), f_\theta(x_i))  \quad \\
    \text{s.t.} & \quad \|\delta_i\|_{\infty} \leq \epsilon \quad \forall i,
\end{split}     \label{eq:trades}
\end{equation}
where $\lambda$ controls the trade-off between robustness and natural accuracy. We follow \cite{zhang2019theoretically} for training algorithms and parameter settings in our experiments.

\begin{table*}[tbhp]
    \centering
    \caption{Performance of (robust) CIFAR-10 models. We inject adaptive layers in WRN-28-4, and compare with WRN-28-4 and WRN-28-5 with more parameters. We provide stronger baselines with our adversarial training ``tricks'' in row 5-7.} 
    \begin{tabular}{|c|c||c|c|c||c|}
    \hline
      \multirow{2}{*}{Row \#}  & \multirow{2}{*}{(Robust) model} & \multicolumn{3}{c||}{ Evaluated Against} & \multirow{2}{*}{\tabincell{c}{\#Parameter\\(million)}} \\ 
     \cline{3-5} & & Natural  & PGD-20 & PGD-100 & \\ 
    \hline\hline
    1 & Natural & \textbf{94.10\%} & 0.00\% & 0.00\% & 5.85\\
    \hline\hline
    2 & PGD-7 \cite{madry2017towards} & 83.84\% & 40.03\% & 39.38\% & 5.85\\
    3 & Free-10 \cite{shafahi2018free} & 81.04\% & 40.56\% & 40.03\% & 5.85\\
    4 & Free-10-adaptive & 85.00\% & 43.16\% & 42.68\% & 6.05\\ 
    \hline \hline
    5 & Free-10-lstep & 77.75\% & 45.10\% & 44.77\% & 5.85\\
    6 & Free-10-WRN-28-5  & 77.81\% & 45.99\% & 45.77\% & 9.13 \\
    7 & Free-10-init  & 80.60\% & 46.88\% & 46.67\% & 5.85 \\
    8 & Free-10-adaptive & 80.99\% & \textbf{48.09\%} & \textbf{47.87\%} & 6.05 \\
    \hline
    \end{tabular}
    \label{tab:c10}
\end{table*}

\section{Experiments}
\label{sec:exp}
In this section, we train robust models on CIFAR-10 and CIFAR-100. In all the experiments, we train WRNs without dropout for 120 epochs and with minibatch size 256. We start with learning rate 0.1 and decrease the learning rate by a factor of 10 at epochs 60 and 90. We use weight decay 1e-4 and momentum 0.9. For evaluating the robustness of the models, we attack them with PGD-K attacks. For the PGD attacks, we use $\epsilon_s=2$ and $\epsilon=8$, and vary the number of attack iterations $K$. 

\subsection{Quantitative evaluation and ablation study}

We summarize our quantitative evaluation on CIFAR-10 and CIFAR-100 in \cref{tab:c10} - \ref{tab:c10large}. In \cref{tab:c10} - \ref{tab:c10fgsm}, unless otherwise explicitly specified through the name of the model, the architecture used for producing these results is WRN-28-4. We report validation accuracy on natural images and adversarial images generated using PGD attacks with $K=20$ iterations and $K=100$ iterations.
We also compare our method with adversarially trained robust models following \cite{madry2017towards} and \cite{shafahi2018free}.
Note that the PGD-7 adversarially trained model~\cite{madry2017towards} requires $\approx 7\times$ more training time than natural training on clean images, while the Free-10 models \cite{shafahi2018free} have similar computation cost as natural training. Models with the suffix ``small'' are adversarially trained using a step-size of $\epsilon_s=2$. The adversarially trained models without the small suffix are trained with a step-size $\epsilon_s=6$. 

\begin{table*}[tbhp]
    \centering
    \caption{Performance of (robust) CIFAR-100 models. Adaptive networks with our adversarial training ``tricks'' in row 8 has 7\% robust accuracy improvement over PGD-7 \cite{madry2017towards} in row 2.} 
    \begin{tabular}{|c|c|c|c|c|c|}
    \hline
      \multirow{2}{*}{Row \#}  & \multirow{2}{*}{(Robust) model} & \multicolumn{3}{c||}{ Evaluated Against} & \multirow{2}{*}{\tabincell{c}{\#Parameter\\(million)}} \\ 
     \cline{3-5} & & Natural  & PGD-20 & PGD-100 & \\ 
    \hline\hline
    1 & Natural & \textbf{74.84\%} & 0.00\% & 0.00\% & 5.87\\
    \hline\hline
    2 & PGD-7 \cite{madry2017towards} & 57.18\%  & 18.38\% & 18.13\% & 5.87\\
    3 &  Free-10 \cite{shafahi2018free} & 54.18\% & 19.21\% & 18.98\% & 5.87\\
    4 & Free-10-adaptive & 61.19\% & 21.95\% & 21.68\% & 6.07\\ 
    \hline \hline
    5 & Free-10-lstep & 50.52\% & 23.08\% & 23.02\% &
    5.87\\
    6 & Free-10-WRN-28-5  & 51.02\% & 23.12\% &  23.03\% & 9.16 \\
    7 & Free-10-init  & 55.93\% & 24.86\% &  24.61\% & 5.87 \\
    8 &  Free-10-adaptive & 57.26\% & \textbf{25.86\%} &  \textbf{25.69\%} & 6.07 \\
    \hline
    \end{tabular}
    \label{tab:c100}
    \vspace{-0.2cm}
\end{table*}

\vspace{0.2cm}
\noindent\textbf{Advantage of adaptive network.} We first evaluate robust models trained with step-size $\epsilon_s=2$ for perturbation updates following \cite{madry2017towards} (rows 2-4 in \cref{tab:c10,tab:c100}). We can train a robust WRN-28-4 with PGD-7 \cite{madry2017towards} that achieves about 40\% accuracy under strong PGD attacks. Our alternative adversarial training mechanism, Free-10 \cite{shafahi2018free} achieves slightly better robust accuracy under PGD attacks with a drop in natural accuracy on clean validation images. Since Free-10 is significantly faster than PGD adversarial training, we also use it to adversarially train our adaptive networks. Our adaptive network with conditional normalization built on WRN-28-4 (Free-10-adaptive, row 4) outperforms the PGD adversarially trained WRN-28-4 (PGD-7, row 2) and Free-10 (row 3) in both natural accuracy and robust accuracy, illustrating the advantage of our adaptive networks.

\vspace{0.2cm}
\noindent\textbf{Strong baseline and effectiveness of our ``tricks'' in adversarial training.} 
We explore ``tricks'' to improve the performance of adversarial training. As shown in \cref{tab:c10,tab:c100},  by comparing Free-10 (row 3) and Free-10-lstep (row 5), we can see that the larger step-size used for training does improve the robustness of free training but again at an additional cost of decreasing natural validation accuracy.

Note that our Free-10-adaptive model has slightly more parameters compared to the adversarially trained PGD-7 and Free-10 models.  For this reason, we compare to higher capacity models to ensure that the superiority of our adaptive network is not solely due to having a (slightly) larger number of parameters.
To create strong, high-capacity baselines we adversarially train a larger model WRN-28-5 (row 6), and WRN-28-4 with a naturally trained model as initialization (row 7).
Our adaptive network is slightly larger than the non-adaptive WRN-28-4, and is much smaller than WRN-28-5.
A good initialization surprisingly helps both natural accuracy and robust accuracy.
Our adaptive network outperforms the best strong baseline for both natural accuracy and robust accuracy.  

\begin{table}[t]
    \centering
    \caption{Performance of (robust) CIFAR-10 WRN-28-4 models with TRADES training \cite{zhang2019theoretically}.
    } 
    \begin{tabular}{|c|c|c|c|}
    \hline
     \multirow{2}{*}{(Robust) model} & \multicolumn{3}{c|}{ Evaluated Against} \\ 
     \cline{2-4} & Natural  & PGD-20 & PGD-100  \\ 
    \hline\hline
    Natural & \textbf{94.90\%} & 0.00\% & 0.00\% \\
    \hline
    Non-adaptive & 84.44\% & 53.74\% & 53.18\%\\
   Adaptive & 84.79	\% &	\textbf{54.98\%} & \textbf{54.76 \%} \\
    \hline
    \end{tabular}
    \label{tab:c10trades}
    \vspace{-0.2cm}
\end{table}

\begin{table}[thbp]
    \centering
    \caption{Performance of (robust) CIFAR-10 WRN-28-4 models with RFGSM training.
    } 
    \vspace{-0.2cm}
    \begin{tabular}{|c|c|c|c|}
    \hline
     \multirow{2}{*}{(Robust) model} & \multicolumn{3}{c|}{ Evaluated Against} \\ 
     \cline{2-4} & Natural  & PGD-20 & PGD-100  \\ 
    \hline\hline
    Natural & \textbf{94.10\%} & 0.00\% & 0.00\% \\
    PGD-7 & 83.84\% & \textbf{40.03\%} & \textbf{39.38\%} \\
    \hline
    RFGSM & 85.81\% & 0.11\% & 0.00\% \\
    Our RFGSM &  84.03 \% & 38.71\% & 37.99 \% \\
    \hline
   Adaptive & 84.87\% & 39.95\% & 38.92\% \\
    \hline
    \end{tabular}
    \label{tab:c10fgsm}
    \vspace{-0.2cm}
\end{table}

\vspace{0.2cm}
\noindent\textbf{TRADES objective and higher robustness.} In \cref{tab:c10trades}, we combine the proposed method with the TRADES objective~\cite{zhang2019theoretically} since our adaptive network is complementary to the objective design of adversarial training. 
We can achieve better robust accuracy on the CIFAR-10 dataset with the TRADES objective, and our adaptive network performs better than the non-adaptive network. 
Note that the TRADES method applies PGD-10 to generate adversarial examples in adversarial training, which is slower than PGD-7 in \cite{madry2017towards}, and much slower than the fast algorithm we used.

\vspace{0.2cm}
\noindent\textbf{RFGSM and the proposed variant.} We present experimental results on RFGSM adversarial training in \cref{tab:c10fgsm}. 
We halved the number of epochs for training so that RFGSM training can complete in similar time as natural training and free adversarial training. 
Classical RFGSM with uniform sampling and norm clipping cannot provide robustness against strong PGD attacks. 
The proposed RFGSM variant can defend against PGD attack, and achieves comparable robust accuracy as PGD adversarial training when combined with our adaptive network.
Though our RFGSM results are worse than our best robust accuracy in \cref{tab:c10} when we use fast adversarial training with ``tricks'' to train the adaptive network,
the proposed variant works well with standard training of ResNet on CIFAR ,
which is complementary to the recent interest in replacing PGD with RFGSM training.

\begin{table}[tbhp]
    \centering
    \caption{Performance of (robust) CIFAR-10 WRN-34-10 models. We directly compare with previously reported results in \cite{madry2017towards,shafahi2018free} and our strong baselines.
    } 
    \setlength{\tabcolsep}{3pt}
    \begin{tabular}{|c|c|c|c|c|}
    \hline
    \multirow{2}{*}{(Robust) model} & \multicolumn{3}{c|}{ Evaluated Against} & \multirow{2}{*}{\tabincell{c}{\#Param\\(million)}} \\ 
     \cline{2-4} & Natural  & PGD-20 & PGD-100 & \\ 
    \hline\hline
     Natural & \textbf{94.76\%} & 0.00\% & 0.00\% & 46.16\\
    \hline
    PGD-7  \cite{madry2017towards} & 87.3\% & 45.8\% & 45.3\% & 45.90\\
    Free-8  \cite{shafahi2018free} & 85.96\% & 46.82\% & 46.19\% & 45.90\\
    \hline 
    Free-10  & 79.45\% &	48.03\% & 47.9 \% & 46.16\\
    Free-10-init  & 84.03\% &	50.23\% &	49.93\% &	46.16\\
    Free-10-adaptive & 84.39\% &	\textbf{50.93\%} &	\textbf{50.68\%} &	47.28  \\
    \hline
    \end{tabular}
    \label{tab:c10large}
\end{table}

\subsection{Larger network and previous benchmark}  
In \cref{tab:c10large}, we report results on a larger network WRN-34-10, which is widely used for the CIFAR-10 benchmark. We first directly compare with the accuracy values reported in the literature in \cite{madry2017towards} and \cite{shafahi2018free} by training on the objective equation \eqref{eq:madry_adv_train}. Our adaptive network achieves better robust accuracy with more than 3\% improvement. Moreover, our adaptive network outperforms the strong baselines we achieved with ``tricks'' (Free-10 and Free-10-init) on both natural accuracy and robust accuracy.

\begin{figure*}[tbhp]
    \centering
    \begin{subfigure}{0.31\linewidth}
        \includegraphics[width=\linewidth]{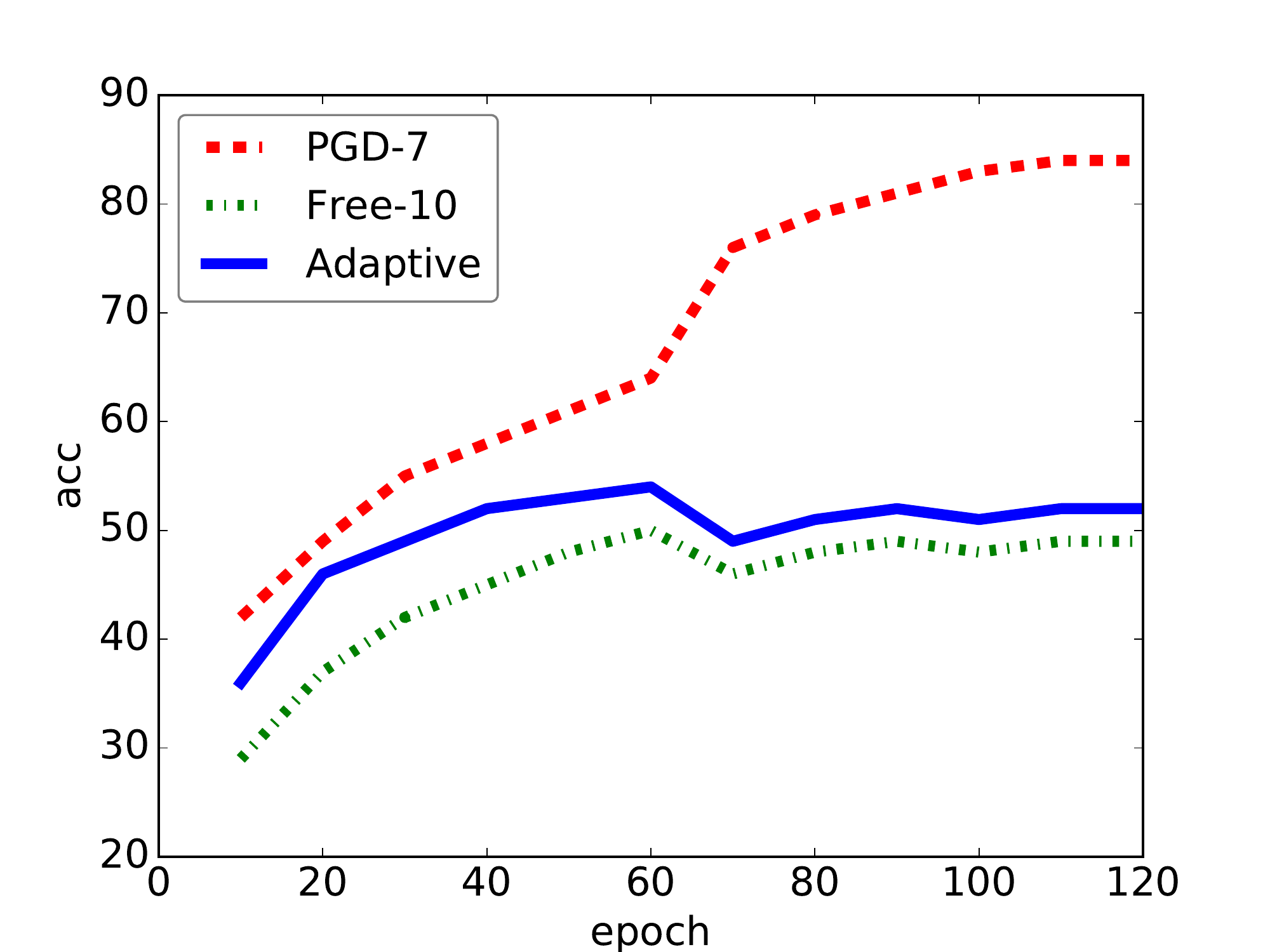}
        \caption{Training accuracy.}
        \label{fig:c10tr}
    \end{subfigure}
    \begin{subfigure}{0.31\linewidth}
        \includegraphics[width=\linewidth]{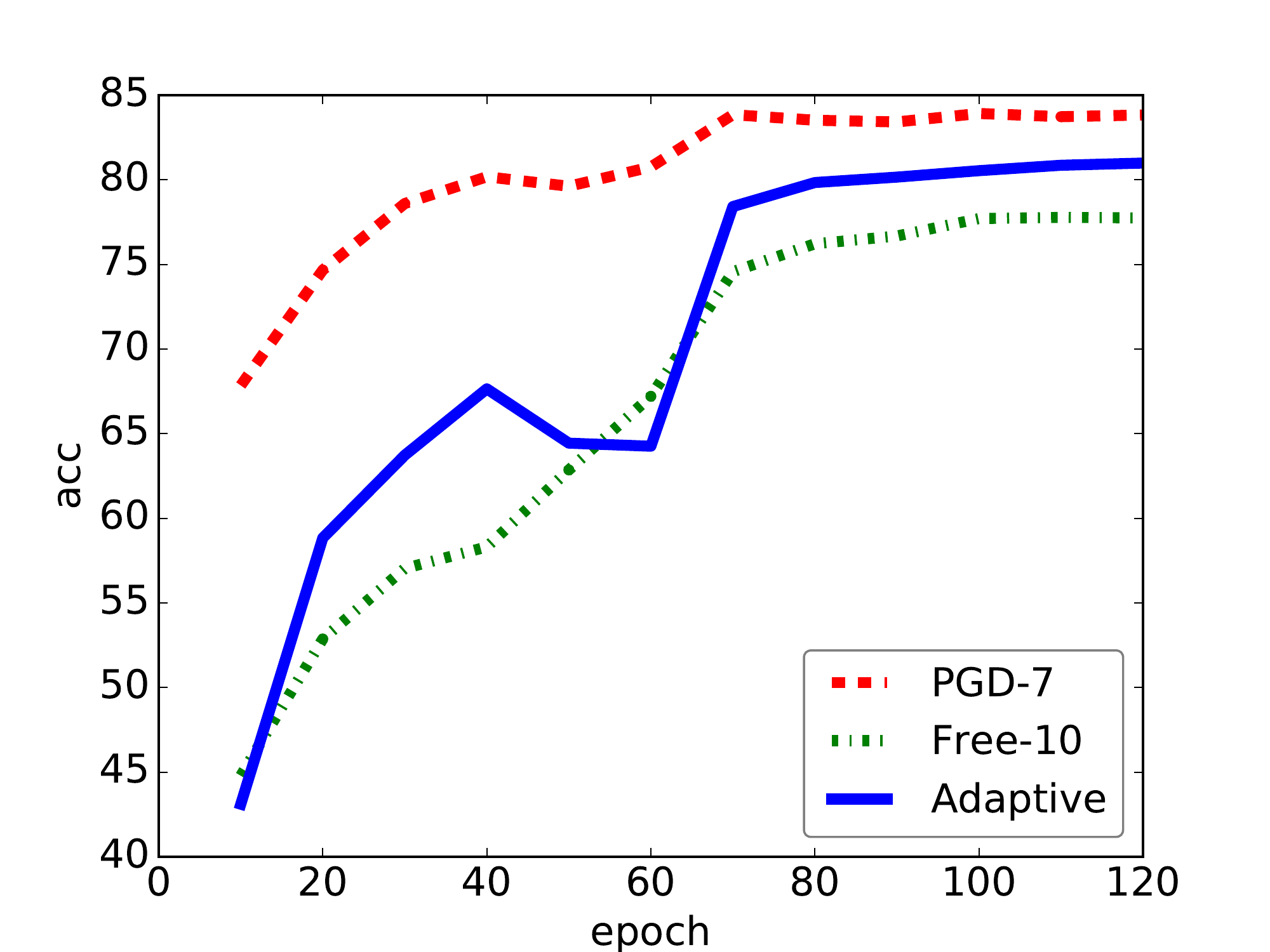}
        \caption{Natural validation accuracy.}
        \label{fig:c10val}    
    \end{subfigure}
    \begin{subfigure}{0.31\linewidth}
        \includegraphics[width=\linewidth]{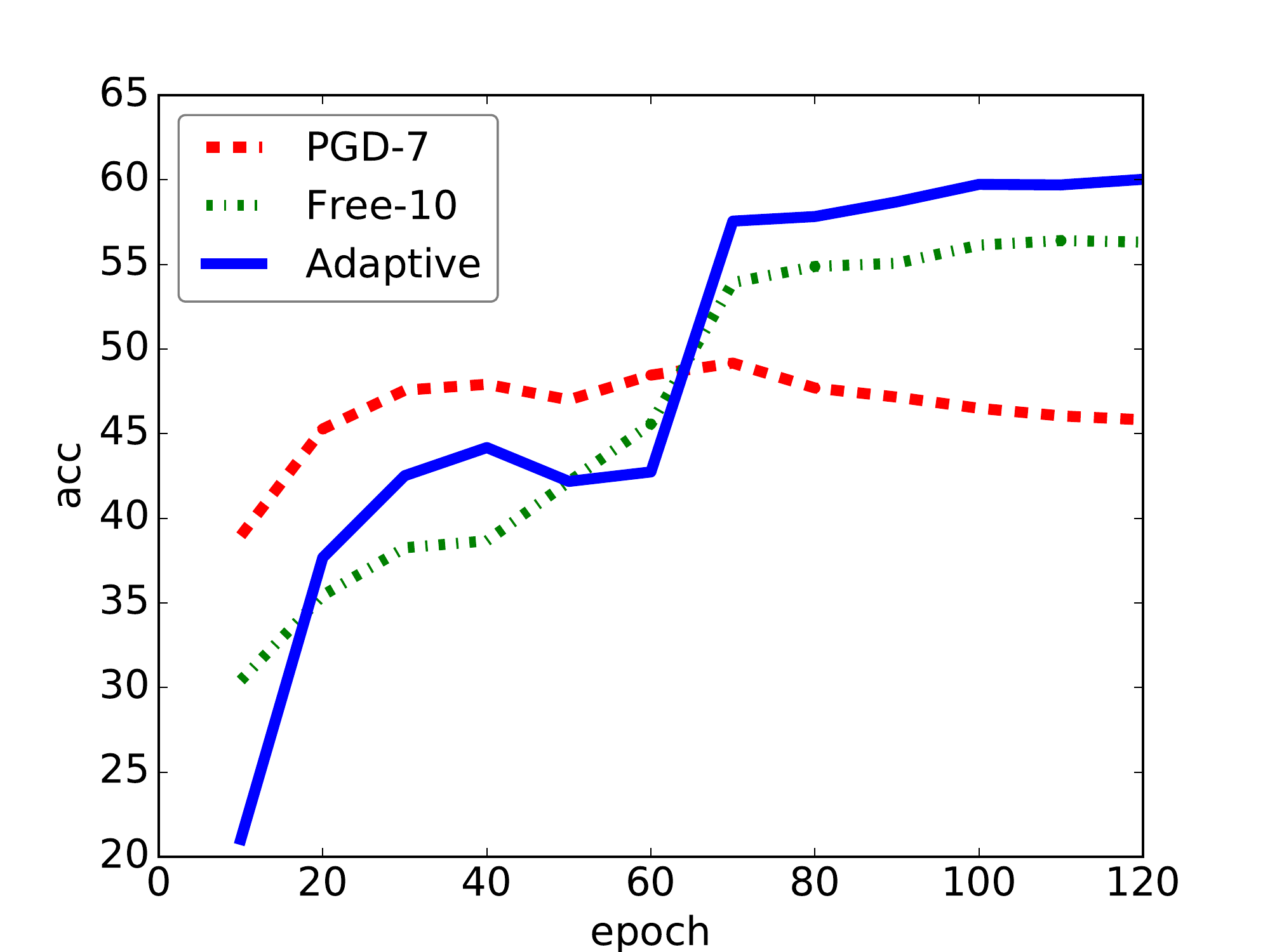}
        \caption{PGD-3 validation accuracy.}
        \label{fig:c10adv}    
    \end{subfigure}
    \begin{subfigure}{0.31\linewidth}
        \includegraphics[width=\linewidth]{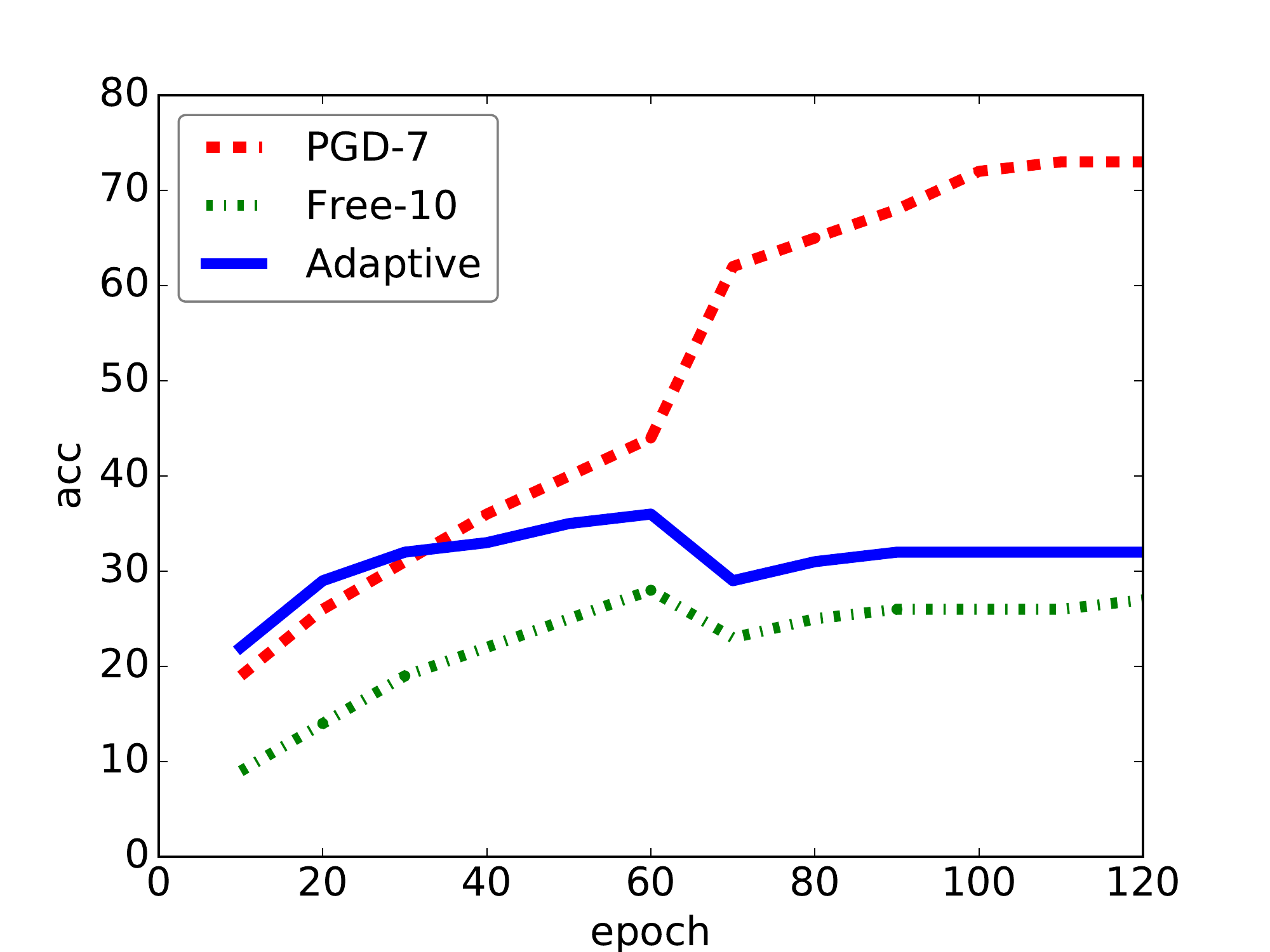}
        \caption{Training accuracy.}
        \label{fig:tr}
    \end{subfigure}
    \begin{subfigure}{0.31\linewidth}
        \includegraphics[width=\linewidth]{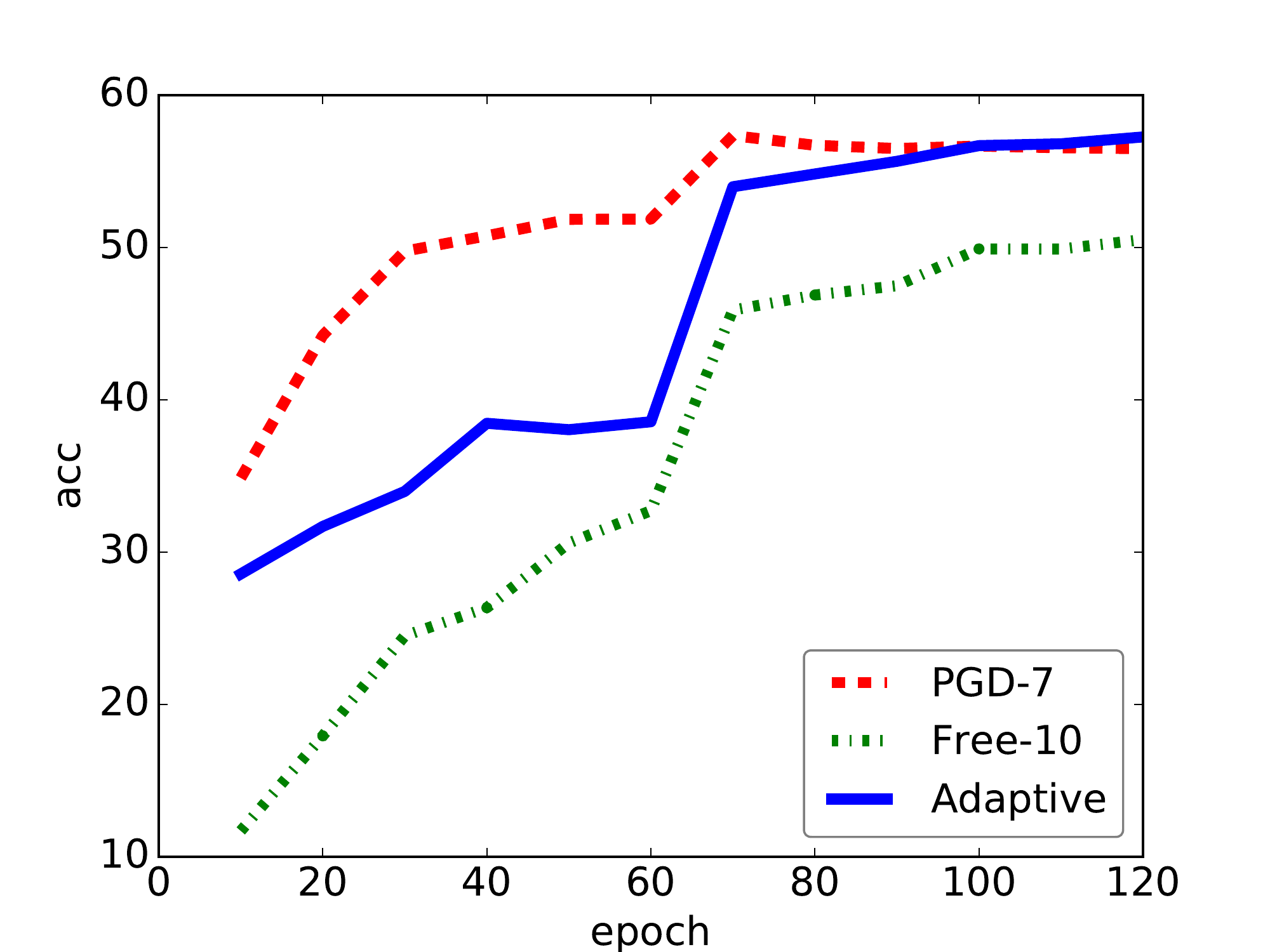}
        \caption{Natural validation accuracy.}
        \label{fig:val}    
    \end{subfigure}
    \begin{subfigure}{0.31\linewidth}
        \includegraphics[width=\linewidth]{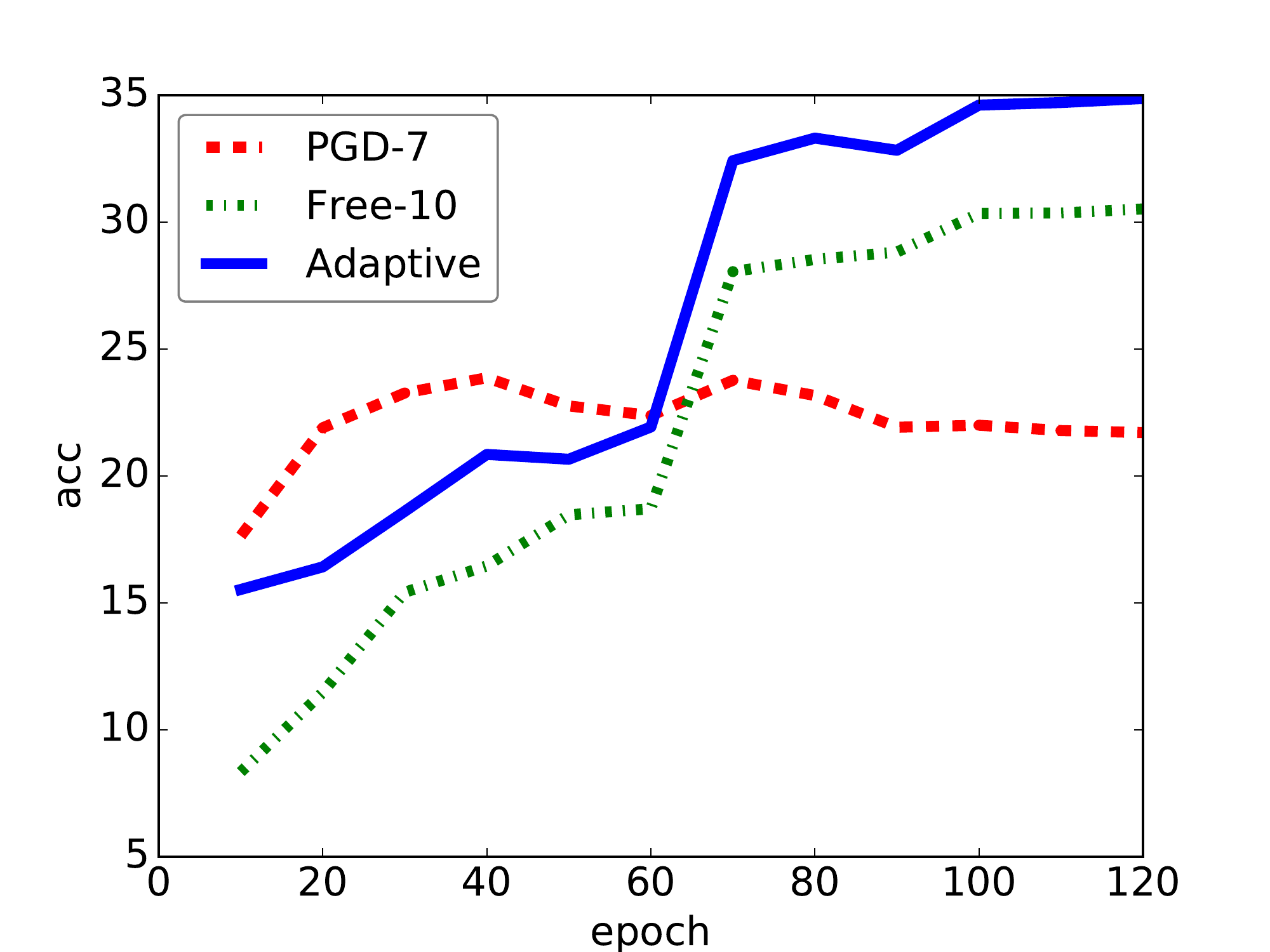}
        \caption{PGD-3 validation accuracy.}
        \label{fig:adv}    
    \end{subfigure}
    \caption{Training curves for robust models for (top) CIFAR-10 and (bottom) CIFAR-100: (left) accuracy on adversarial training samples; (middle) accuracy on clean validation samples; (right) accuracy on PGD-3 validation samples.}
    \label{fig:curve}
    \vspace{-0.2cm}
\end{figure*}

\begin{figure*}[thbp]
    \centering
    \includegraphics[width=0.98\linewidth]{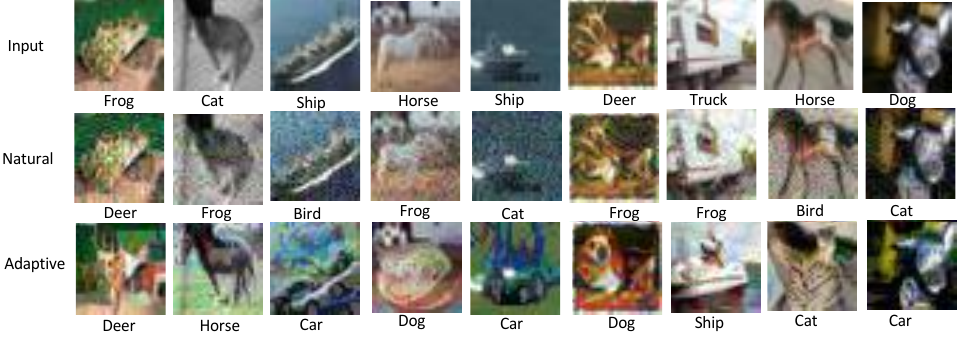}
    \caption{
    Visualization of adversarial examples generated for natural
 and robust WRN-34-10 for CIFAR-10 with large $\epsilon=30$ following \cite{tsipras2018robustness}. 
 The large $\epsilon$ adversarial examples generated for
 robust models align well with human perception.
    }
    \label{fig:vis}
\end{figure*}

\subsection{Training curves and qualitative analysis}
We plot the training and validation accuracy of the Free-10, Free-10-adaptive, and PGD-7 adversarially trained (PGD-7) models after each epoch in \cref{fig:curve}. Training accuracy of robust models is computed using adversarial examples that are seen during training, and do not correspond to the natural training accuracy. They can be thought of as robustness on training examples. In \cref{fig:c10tr,fig:tr}, the PGD-7 model fits the adversarial examples built for the training samples to a rather high accuracy, while Free-10 seems to never overfit to the training-set adversarial training samples. The training accuracy of Free-10 \cite{shafahi2018free} is quite close to the final adversarial validation accuracy in \cref{fig:c10adv,fig:adv}. 
The natural validation accuracy of PGD-7 increases faster than Free-10 at the beginning, while the accuracy at the end of training become close, as shown in \cref{fig:c10val,fig:val}. 
Free-10 consistently improves robust accuracy against adversarial validation samples, while PGD-7 seems to saturate after the fast increase at the beginning (see \cref{fig:c10adv,fig:adv}). 
Our adaptive network (blue curve) always has higher natural and robust validation accuracy than the non-adaptive WRN-28-4 models except for a short range around epoch 60 in \cref{fig:c10val,fig:c10adv}, where the accuracy of the adaptive network decreases. Tuning the learning rate could potentially prevent this decrease and further boost the performance of adaptive networks.

\citet{tsipras2018robustness} presented an interesting side effect of robust models: largely perturbed adversarial examples for adversarially robust models align with human perception. That is, they ``look'' like the class which they are getting misclassified to. We use PGD-50 to generate adversarial images with large perturbations ($\epsilon=30$). The generated images for our adversarially trained adaptive nets have characteristics that align well with human perception (Fig.~\ref{fig:vis}).

\section{Conclusion}
Inspired by recent research in conditional normalization \cite{huang2017arbitrary,karras2018style} and properties of robustness \cite{madry2017towards,tsipras2018robustness,schmidt2018adversarially},
we introduced an adaptive normalization module conditioned on inputs for boosting the robustness of networks. 
Our adaptive networks combined with a fast adversarial training algorithm, can effectively train robust models that outperform their non-adaptive parallels and also non-adaptive networks with more parameters. 
Our study on adversarial training presents several ``tricks'' that can be widely used to improve the training performance of robust models. 
We also introduce a variant of singe-step adversarial training that can achieve competitive robustness against multi-step attacks.
We verify the effectiveness and efficiency of adaptive networks and our adversarial training with experiments on CIFAR-10 and CIFAR-100 benchmark and WRN networks.

\bibliographystyle{abbrvnat}
\bibliography{adversarial,style,gan}

\end{document}